\documentclass[11pt]{article}

\usepackage[preprint]{acl}
\usepackage{times}
\usepackage{latexsym}
\usepackage[T1]{fontenc}
\usepackage[utf8]{inputenc}
\usepackage{microtype}
\usepackage{inconsolata}
\usepackage{graphicx}
\usepackage{booktabs}
\usepackage{amsmath}
\usepackage{amssymb}
\usepackage{xcolor}
\usepackage{multirow}
\usepackage{url}
\usepackage{algorithm}
\usepackage{algpseudocode}
\usepackage{tikz}
\usetikzlibrary{arrows.meta,backgrounds,calc,fit,positioning,shapes.geometric}

\newcommand{\method}{\textsc{SRR}}
\newcommand{\smallmodel}{Qwen3-4B}
\newcommand{\largemodel}{Qwen3-8B}
\newcommand{\NumExamples}{5,165}
\newcommand{\SmallAccuracy}{61.8}
\newcommand{\LargeAccuracy}{68.9}
\newcommand{\SRRAUACC}{0.6811}
\newcommand{\ErrorAUACC}{0.6716}
\newcommand{\EntropyAUACC}{0.6659}
\newcommand{\SRRAtTwentyFive}{67.6}
\newcommand{\ErrorAtTwentyFive}{66.1}
\newcommand{\SRRAtFifty}{69.4}
\newcommand{\ErrorAtFifty}{68.1}
\newcommand{\SmallLatency}{12.80}
\newcommand{\LargeLatency}{19.60}
\newcommand{\ResultRescueRate}{12.4\%}
\newcommand{\ResultHarmRate}{5.3\%}
\newcommand{\ResultRandomAUACC}{0.6534}
\newcommand{\ResultRandomAtTwentyFive}{63.5}
\newcommand{\ResultRandomAtFifty}{65.4}
\newcommand{\ResultConfidenceAUACC}{0.6646}
\newcommand{\ResultConfidenceAtTwentyFive}{65.0}
\newcommand{\ResultConfidenceAtFifty}{67.1}

\newcommand{\ResultEntropyAtTwentyFive}{65.2}
\newcommand{\ResultEntropyAtFifty}{67.3}
\newcommand{\ResultRescueAUACC}{0.6751}
\newcommand{\ResultRescueAtTwentyFive}{66.5}
\newcommand{\ResultRescueAtFifty}{68.7}
\newcommand{\ResultOracleAUACC}{0.7121}
\newcommand{\ResultOracleAtTwentyFive}{71.1}
\newcommand{\ResultOracleAtFifty}{74.2}

\title{Signed Rescue Routing: Harm-Aware Cascades for Efficient LLM Inference}

\author{
Zheyuan Wang$^{1}$,
Siyu Li$^{2}$,
Peiqiao Song$^{3}$,
Sijia Chen$^{4}$,
Qianqian Song$^{1}$,
Qiao Liu$^{1}$\\
$^{1}$School of Artificial Intelligence, Beijing Normal University\\
$^{2}$School of Information Science and Engineering, Chongqing Jiaotong University\\
$^{3}$School of Computer Science and Technology, Dalian University of Technology\\
$^{4}$School of Computer Science and Technology, Jilin University
}

\begin{document}
\maketitle

\begin{abstract}
Large language model (LLM) cascades answer easy requests with a small model
and escalate selected requests to a larger model.
Most routers prioritize examples on which the small model appears uncertain
or likely to be wrong.
This proxy ignores a decisive fact: escalation is useful only when the large
model corrects the small model, and it is harmful when the large model
replaces a correct answer with an incorrect one.
We introduce \emph{Signed Rescue Routing} (\method), a budgeted routing
method that predicts these two events separately and ranks requests by
their difference.
We show that this signed conditional gain is the Bayes-optimal routing score
under a fixed escalation budget.
\method{} requires only the small model's output statistics at deployment and
adds a lightweight two-head router.
We evaluate \method{} with \smallmodel{} and \largemodel{} on
\NumExamples{} examples from MMLU, HellaSwag, and ARC-Challenge.
Across the accuracy--compute curve, \method{} reaches an area of
\SRRAUACC{}, compared with \ErrorAUACC{} for a learned small-model error
predictor and \EntropyAUACC{} for entropy routing.
These results show that predicting \emph{incremental value}, rather than
model uncertainty, is a simple and effective objective for efficient LLM
cascades.
\end{abstract}

\section{Introduction}

Serving every request with the strongest available language model is often
unnecessary.
Model cascades reduce inference cost by first running a smaller model and
escalating only selected requests to a larger model
\citep{chen2023frugalgpt,ong2024routellm,ding2024hybrid}.
The central problem is therefore not only whether the small model is
uncertain, but whether paying for the larger model changes the outcome for
the better.

Existing routing rules commonly use confidence, entropy, self-consistency,
or a learned prediction that the small model is wrong
\citep{shnitzer2023routing,ong2024routellm}.
These signals can be effective, but they optimize a proxy.
An uncertain small-model answer need not be repairable by the large model;
conversely, the large model can overturn a correct small-model answer.
Both cases spend additional compute without improving accuracy, and the
second strictly reduces it.
The distinction is especially important when model errors are correlated or
when the larger model is only moderately more accurate.

We propose \emph{Signed Rescue Routing} (\method).
For each request $x$, let $Y_s,Y_\ell\in\{0,1\}$ denote whether the small
and large models are correct.
\method{} predicts
\begin{equation}
  \begin{aligned}
  G(x)={}&\Pr(Y_s=0,Y_\ell=1\mid z_x)\\
         &-\Pr(Y_s=1,Y_\ell=0\mid z_x),
  \end{aligned}
  \label{eq:signed-gain}
\end{equation}
where $z_x$ contains features available after the small-model call.
The first term is the probability of a \emph{rescue}; the second is the
probability of \emph{harm}.
Under a fixed escalation budget, routing the requests with the largest
$G(x)$ maximizes expected cascade accuracy.

We predict the two rare events with separate cost-sensitive heads.
The runtime feature vector uses only answer probabilities, margin, entropy,
prompt length, and categorical task/output features from the small model.
No large-model computation is needed unless the router chooses to escalate.
This design makes the routing objective explicit and keeps the deployment
path lightweight.

Our contributions are:
\begin{itemize}
  \item We derive signed conditional gain as the Bayes-optimal score for
  budgeted LLM cascades and identify confidence and small-model error
  prediction as incomplete surrogates.
  \item We introduce \method, a practical two-head predictor that separately
  models rescue and harm events using only small-model-side features.
  \item We conduct a controlled evaluation on three public multiple-choice
  benchmarks using open Qwen3 models, reporting complete accuracy--budget
  curves and measured P4 latency.
\end{itemize}

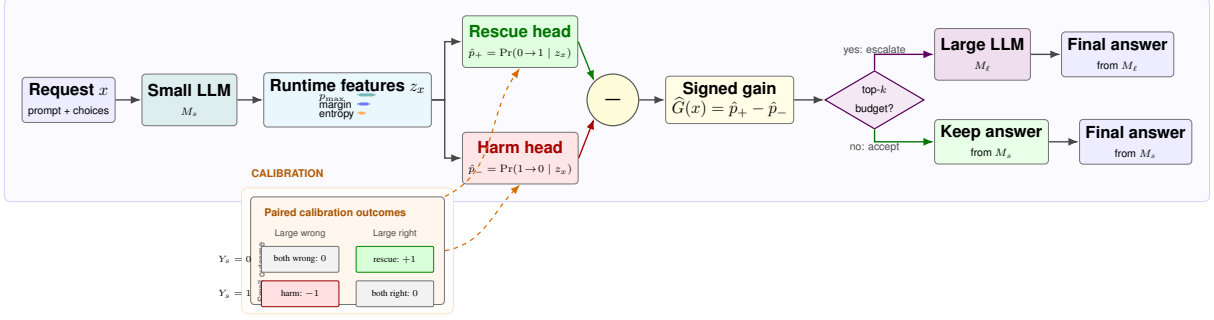
\begin{figure*}[t]
  \centering
  \resizebox{\textwidth}{!}{\begin{tikzpicture}[
  font=\sffamily,
  node distance=5mm and 6mm,
  >=Latex,
  box/.style={rounded corners=2.5pt, draw=black!55, line width=.65pt,
    minimum height=9mm, align=center, inner sep=4pt},
  stage/.style={box, fill=blue!7},
  model/.style={box, minimum width=21mm, minimum height=12mm},
  head/.style={box, minimum width=23mm, minimum height=12mm},
  flow/.style={->, line width=.9pt, draw=black!72},
  train/.style={->, dashed, line width=.8pt, draw=orange!85!black},
  tinylabel/.style={font=\scriptsize\sffamily, text=black!70},
]

\node[stage, minimum width=20mm] (query) {
  \textbf{Request $x$}\\[-1pt]
  \scriptsize prompt + choices
};
\node[model, fill=teal!13, right=of query] (small) {
  \textbf{Small LLM}\\[-1pt]
  \scriptsize $M_s$
};
\node[box, fill=cyan!7, minimum width=29mm, right=of small] (features) {
  \textbf{Runtime features $z_x$}\\[-2pt]
  \scriptsize
  \begin{tabular}{@{}l@{\hspace{2pt}}l@{}}
    $p_{\max}$ & \tikz[baseline=-.5ex]\fill[teal!55] (0,0) rectangle (0.43,.07);\\[-2pt]
    margin & \tikz[baseline=-.5ex]\fill[blue!55] (0,0) rectangle (0.30,.07);\\[-2pt]
    entropy & \tikz[baseline=-.5ex]\fill[orange!65] (0,0) rectangle (0.20,.07);
  \end{tabular}
};

\node[head, fill=green!12, above right=1mm and 7mm of features] (rescue) {
  \textbf{\color{green!45!black}Rescue head}\\[-1pt]
  \scriptsize $\hat p_{+}=\Pr(0\!\to\!1\mid z_x)$
};
\node[head, fill=red!10, below right=1mm and 7mm of features] (harm) {
  \textbf{\color{red!65!black}Harm head}\\[-1pt]
  \scriptsize $\hat p_{-}=\Pr(1\!\to\!0\mid z_x)$
};
\coordinate (headmid) at ($(rescue.east)!0.5!(harm.east)$);
\node[circle, draw=black!65, fill=yellow!18, line width=.8pt,
  minimum size=12mm] (subtract) at ($(headmid)+(8mm,0)$) {
  \Large$\boldsymbol{-}$
};
\node[box, fill=yellow!12, minimum width=24mm, right=of subtract] (score) {
  \textbf{Signed gain}\\[-1pt]
  $\widehat G(x)=\hat p_+-\hat p_-$
};
\node[diamond, aspect=1.7, draw=violet!70!black, fill=violet!9,
  line width=.8pt, align=center, inner sep=1.5pt, right=7mm of score] (gate) {
  \scriptsize top-$k$\\[-1pt]\scriptsize budget?
};
\node[model, fill=violet!13, above right=1mm and 8mm of gate] (large) {
  \textbf{Large LLM}\\[-1pt]
  \scriptsize $M_\ell$
};
\node[box, fill=green!9, below right=1mm and 8mm of gate,
  minimum width=23mm] (keep) {
  \textbf{Keep answer}\\[-1pt]
  \scriptsize from $M_s$
};
\node[box, fill=blue!7, right=7mm of large, minimum width=21mm] (finalbig) {
  \textbf{Final answer}\\[-1pt]
  \scriptsize from $M_\ell$
};
\node[box, fill=blue!7, right=7mm of keep, minimum width=21mm] (finalsmall) {
  \textbf{Final answer}\\[-1pt]
  \scriptsize from $M_s$
};

\draw[flow] (query) -- (small);
\draw[flow] (small) -- (features);
\draw[flow] (features.east) -- ++(3mm,0) |- (rescue.west);
\draw[flow] (features.east) -- ++(3mm,0) |- (harm.west);
\draw[flow, draw=green!50!black] (rescue.east) -- (subtract.north west);
\draw[flow, draw=red!65!black] (harm.east) -- (subtract.south west);
\draw[flow] (subtract) -- (score);
\draw[flow] (score) -- (gate);
\draw[flow, draw=violet!75!black] (gate) |- node[pos=.28,above,tinylabel]{yes: escalate} (large);
\draw[flow, draw=green!45!black] (gate) |- node[pos=.28,below,tinylabel]{no: accept} (keep);
\draw[flow] (large) -- (finalbig);
\draw[flow] (keep) -- (finalsmall);

\node[box, fill=orange!7, minimum width=45mm, minimum height=25mm,
  below=16mm of features, anchor=north] (matrix) {};
\node[anchor=north west, font=\scriptsize\sffamily\bfseries,
  text=orange!65!black] at ([xshift=2mm,yshift=-1.5mm]matrix.north west)
  {Paired calibration outcomes};
\node[font=\tiny\sffamily, text=black!70] at ([xshift=-11mm,yshift=4mm]matrix.center)
  {Large wrong};
\node[font=\tiny\sffamily, text=black!70] at ([xshift=11mm,yshift=4mm]matrix.center)
  {Large right};
\node[font=\tiny\sffamily, rotate=90, text=black!70]
  at ([xshift=-20mm,yshift=-5mm]matrix.center) {Small outcome};
\node[box, rounded corners=1pt, fill=gray!10, minimum width=18mm,
  minimum height=6mm, font=\tiny] at ([xshift=-11mm,yshift=-2mm]matrix.center)
  {both wrong: $0$};
\node[box, rounded corners=1pt, fill=green!15, draw=green!55!black,
  minimum width=18mm, minimum height=6mm, font=\tiny]
  at ([xshift=11mm,yshift=-2mm]matrix.center) {rescue: $+1$};
\node[box, rounded corners=1pt, fill=red!12, draw=red!60!black,
  minimum width=18mm, minimum height=6mm, font=\tiny]
  at ([xshift=-11mm,yshift=-10mm]matrix.center) {harm: $-1$};
\node[box, rounded corners=1pt, fill=gray!10, minimum width=18mm,
  minimum height=6mm, font=\tiny] at ([xshift=11mm,yshift=-10mm]matrix.center)
  {both right: $0$};
\node[font=\tiny\sffamily, anchor=east] at ([xshift=-21mm,yshift=-2mm]matrix.center)
  {$Y_s=0$};
\node[font=\tiny\sffamily, anchor=east] at ([xshift=-21mm,yshift=-10mm]matrix.center)
  {$Y_s=1$};

\draw[train] (matrix.north east) to[out=25,in=-125] (rescue.south);
\draw[train] (matrix.east) to[out=10,in=-135] (harm.south);

\begin{scope}[on background layer]
  \node[fit=(query)(small)(features)(rescue)(harm)(subtract)(score)(gate)
    (large)(keep)(finalbig)(finalsmall),
    rounded corners=4pt, fill=blue!1.5, draw=blue!18, inner sep=4mm] (inferband) {};
  \node[fit=(matrix), rounded corners=4pt, fill=orange!1.5,
    draw=orange!22, inner sep=2mm] (trainband) {};
\end{scope}
\node[anchor=south west, font=\scriptsize\sffamily\bfseries,
  text=blue!65!black] at ([xshift=1mm,yshift=1mm]inferband.north west)
  {INFERENCE: predict incremental value before spending large-model compute};
\node[anchor=south west, font=\scriptsize\sffamily\bfseries,
  text=orange!70!black] at ([xshift=1mm,yshift=1mm]trainband.north west)
  {CALIBRATION};

\end{tikzpicture}}
  \caption{\textbf{Signed Rescue Routing overview.}
  The small model answers every request and exposes inexpensive runtime
  features. Two separately supervised heads predict the probability that
  escalation repairs an error (rescue, green) and the probability that it
  overturns a correct answer (harm, red). Their difference is the predicted
  incremental value of calling the large model. A budget gate escalates only
  the highest-value requests. Dashed orange arrows denote calibration-time
  supervision; solid arrows denote the deployment path.}
  \label{fig:method}
\end{figure*}

\section{Why Error Prediction Is Not Enough}
\label{sec:motivation}

The standard cascade intuition is ``defer when the small model is likely
wrong.'' This is correct only when the large model is an error-free oracle.
With two fallible models, paired outcomes form the four cells shown in the
calibration inset of Figure~\ref{fig:method}.
Two cells are neutral: both models are correct or both are wrong.
Only a rescue contributes $+1$ accuracy, while a harm contributes $-1$.
The expected value of escalation is therefore signed.

\paragraph{Correlated errors waste budget.}
If both models fail on the same hard examples, an error router will
concentrate expensive calls exactly where they are least likely to help.
This failure does not disappear with a better prediction of small-model
uncertainty: the missing variable is the conditional capability of the
large model.

\paragraph{Non-monotonic model quality creates harm.}
Larger models are stronger on average, but model quality is not pointwise
monotonic.
Prompt sensitivity, domain specialization, and answer-format behavior can
make the smaller model correct on examples that the larger model misses.
FrugalGPT explicitly reports complementary errors among model pairs
\citep{chen2023frugalgpt}; Hybrid LLM similarly motivates routing from the
tail of per-example quality differences \citep{ding2024hybrid}.
\method{} turns this complementarity into the supervised routing target.

\paragraph{The router predicts a treatment effect.}
Escalation can be viewed as an intervention whose outcome is
$\Delta=Y_\ell-Y_s$.
The router is not asked to predict whether an example is globally hard; it
predicts the conditional effect of changing the serving policy from
$M_s$ to $M_\ell$.
This perspective separates model confidence from model complementarity and
directly matches the cascade objective.

\section{Related Work}

\paragraph{LLM cascades and routing.}
FrugalGPT formulates cost-aware combinations of language model APIs
\citep{chen2023frugalgpt}.
Hybrid LLM and RouteLLM learn when to send queries to a stronger model
\citep{ding2024hybrid,ong2024routellm}, while routing benchmarks study the
relationship between query difficulty and model choice
\citep{shnitzer2023routing}.
AutoMix uses answer verification to decide whether to switch models,
FORC learns cost-aware model choice, and RouterBench standardizes
multi-model evaluation
\citep{madaan2024automix,sakota2024forc,hu2024routerbench}.
Recent work extends routing to multiple models, preference signals, and
routing benchmarks
\citep{lu2023zooter,li2026routerbench,rabanser2025gatekeeper}.
Our work focuses on a complementary issue: the correct supervised target for
a two-model cascade.
Rather than predict absolute correctness or preference, \method{} predicts the
signed outcome change caused by escalation.
The closest conceptual connection is deferral optimization: Gatekeeper
improves confidence predictions so the realized deferral curve approaches an
ideal curve \citep{rabanser2025gatekeeper}.
\method{} instead changes the target being calibrated, from small-model
correctness to the signed benefit of replacing its answer.

\paragraph{Selective prediction.}
Classification with a reject option trades coverage for risk
\citep{chow1970reject,elyaniv2010selective,geifman2019selectivenet}.
Modern neural networks can be miscalibrated, especially under distribution
shift, and selective question answering explicitly studies when a model
should abstain
\citep{guo2017calibration,ovadia2019uncertainty,kamath2020selectiveqa}.
An LLM cascade differs because rejection invokes another fallible predictor.
The downstream model may rescue or harm the answer, so the relevant utility
is the conditional difference between the two outcomes.

\paragraph{Efficient LLM inference.}
Other efficiency techniques reduce the cost of each model call through
weight and activation quantization
\citep{dettmers2022llmint8,frantar2023gptq,xiao2023smoothquant,lin2024awq},
memory-efficient attention
\citep{dao2022flashattention}, dynamic sparse attention for long-context
inference \citep{xionglong}, speculative decoding
\citep{leviathan2023speculative}, or serving optimizations
\citep{sheng2023flexgen,yu2022orca,kwon2023vllm}.
Cascading is orthogonal: it reduces how often the expensive model is called
and can be combined with these methods.

\begin{figure}[t]
  \centering
  \IfFileExists{figures/accuracy_budget.pdf}{
    \includegraphics[width=0.7\columnwidth]{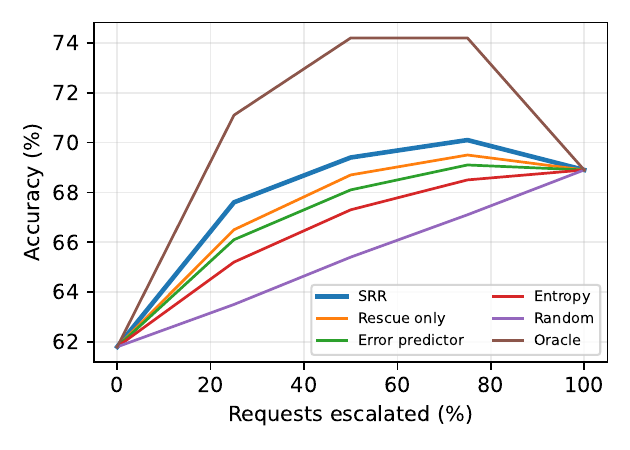}
  }{
    \fbox{\parbox[c][3.4cm][c]{0.92\columnwidth}{
      \centering Accuracy--budget figure generated after experiments.}}
  }
  \caption{Held-out cascade accuracy as a function of the
  fraction of requests escalated to \largemodel{}.}
  \label{fig:curve}
\end{figure}

\paragraph{Visual and evaluation conventions.}
Recent routing papers typically pair a compact system diagram with a
performance--cost curve.
FrugalGPT contrasts standard single-model usage with a learned cascade
\citep{chen2023frugalgpt}; Gatekeeper places the cascade flow beside ideal
and realized deferral curves \citep{rabanser2025gatekeeper}; RouteLLM
emphasizes the area and thresholds on the routing curve
\citep{ong2024routellm}.
We follow this convention in Figures~\ref{fig:method} and
\ref{fig:curve}, while exposing the paired rescue/harm supervision that is
specific to \method.

\section{Problem Formulation}

Let a small model $M_s$ answer every request and a larger model $M_\ell$
answer a routed subset.
For request $x_i$, correctness variables $Y_{s,i}$ and $Y_{\ell,i}$ are
binary.
A router chooses $R_i\in\{0,1\}$, where $R_i=1$ replaces the small-model
answer with the large-model answer.
Cascade accuracy is
\begin{equation}
  A(R)
  =
  \frac{1}{n}\sum_{i=1}^n
  \left[(1-R_i)Y_{s,i}+R_iY_{\ell,i}\right].
  \label{eq:accuracy}
\end{equation}
With a uniform large-model cost, the budget constraint is
$\sum_i R_i\leq k$.
The realized gain from routing example $i$ is
\begin{equation}
  \Delta_i=Y_{\ell,i}-Y_{s,i}\in\{-1,0,1\}.
\end{equation}
The three values correspond to harm, no change, and rescue.

\paragraph{Proposition 1.}
Let $z_i$ be the information available to the router after running $M_s$.
Among all policies that route $k$ of $n$ requests, selecting the $k$
requests with largest
\begin{equation}
  \begin{aligned}
  \mathbb{E}[\Delta_i\mid z_i]
  ={}&\Pr(Y_s=0,Y_\ell=1\mid z_i)\\
     &-\Pr(Y_s=1,Y_\ell=0\mid z_i)
  \end{aligned}
  \label{eq:optimal-score}
\end{equation}
maximizes expected accuracy.

\paragraph{Proof.}
Conditioned on the observed features, Equation~\ref{eq:accuracy} equals the
small-model accuracy plus
$n^{-1}\sum_i R_i\mathbb{E}[\Delta_i\mid z_i]$.
If a selected request has lower expected gain than an unselected request,
swapping them weakly increases the objective.
Repeated exchanges yield the top-$k$ policy. \hfill$\square$

This result exposes the limitation of uncertainty routing.
Predicting $\Pr(Y_s=0\mid z)$ treats all small-model errors as equally
repairable and assigns no negative value to possible harm.
The approximation is justified only under restrictive assumptions, such as
the large model correcting every routed small-model error and never
overturning a correct answer.

\section{Signed Rescue Routing}

\subsection{Two-Head Gain Prediction}

\method{} decomposes Equation~\ref{eq:optimal-score} into two binary events:
\begin{align}
  p_{\mathrm{rescue}}(z)
  &=\Pr(Y_s=0,Y_\ell=1\mid z),\\
  p_{\mathrm{harm}}(z)
  &=\Pr(Y_s=1,Y_\ell=0\mid z).
\end{align}
Two lightweight classifiers are trained on a calibration set containing
paired small- and large-model outcomes.
The routing score is
\begin{equation}
  \widehat G(z)
  =
  \widehat p_{\mathrm{rescue}}(z)
  -
  \widehat p_{\mathrm{harm}}(z).
\end{equation}
Separating the heads preserves the asymmetric semantics of the two rare
events and permits event-specific class weighting.

For calibration example $i$, define
\begin{align}
  r_i &= \mathbb{1}[Y_{s,i}=0 \land Y_{\ell,i}=1],\\
  h_i &= \mathbb{1}[Y_{s,i}=1 \land Y_{\ell,i}=0].
\end{align}
The rescue and harm heads, parameterized by $\theta_+$ and $\theta_-$, are
optimized independently:
\begin{equation}
  \mathcal{L}
  =
  \operatorname{WBCE}(r,\hat p_+)
  +\lambda_h\operatorname{WBCE}(h,\hat p_-),
  \label{eq:loss}
\end{equation}
where WBCE is weighted binary cross-entropy and $\lambda_h$ controls the
relative cost of routing a harmful example.
We use inverse-frequency positive weights for both heads and
$\lambda_h=1$ unless otherwise noted.
Unlike a three-way classifier, this factorization keeps both probabilities
interpretable and permits asymmetric deployment policies.

\subsection{Relation to Alternative Targets}

Several natural targets are special cases or incomplete approximations:
\begin{itemize}
  \item \textbf{Uncertainty routing} ranks by entropy or $1-p_{\max}$ and
  does not use paired large-model outcomes.
  \item \textbf{Error routing} predicts $\Pr(Y_s=0\mid z)$, implicitly
  treating shared failures as rescues and ignoring harm.
  \item \textbf{Rescue-only routing} predicts $\Pr(Y_s=0,Y_\ell=1\mid z)$
  but assigns zero cost to overwriting a correct answer.
  \item \textbf{\method} predicts the full conditional mean of
  $Y_\ell-Y_s$, which is the quantity appearing in the cascade objective.
\end{itemize}
The target ablation in Table~\ref{tab:ablation} compares these choices while
holding features and router capacity fixed.

\subsection{Runtime Features}

For a four-choice request, the small model produces normalized option
probabilities $p\in\mathbb{R}^4$.
We use maximum probability, top-two margin, entropy, prompt length, the
standard deviations of probabilities and log probabilities, the predicted
option identity, and a task identifier.
All features are available from the small-model call.
The deployed router is a two-layer multilayer perceptron with 32 hidden
units; its overhead is negligible relative to either LLM forward pass.
Numeric features are standardized on the calibration split.
Categorical features are one-hot encoded.
The complete feature vector has 13 dimensions in our four-choice setting;
no hidden states, extra generations, or large-model features are required
at runtime.

\subsection{Budgeted Deployment}

For an offline batch, \method{} routes the top $k$ scores.
For an online stream, the same rule can be implemented by calibrating a score
threshold to the desired escalation rate and updating its empirical
quantile.
When request costs vary, Proposition~1 generalizes to a knapsack objective;
the uniform-cost setting isolates the routing signal studied here.

\paragraph{Cost-sensitive extension.}
If escalation cost $c_i$ differs across requests---for example because prompt
lengths vary---the optimal unconstrained Lagrangian policy routes when
\begin{equation}
  \widehat G(z_i)-\eta c_i>0,
  \label{eq:cost-sensitive}
\end{equation}
for multiplier $\eta$ selected to meet the aggregate budget.
Ranking by $\widehat G/c_i$ gives the usual fractional-knapsack relaxation.
We retain uniform cost in the main experiment so every router receives
exactly the same number of large-model calls.

\paragraph{Online serving procedure.}
The deployed system performs four steps: (1) execute $M_s$ and extract
$z_x$; (2) evaluate the two router heads; (3) compare the signed score with
the budget-controlled quantile threshold; and (4) either keep the small
answer or replace it with $M_\ell$'s answer.
The heads add $O(dH)$ operations for feature dimension $d$ and hidden width
$H$, versus billions of operations for an LLM call.
Algorithm~\ref{alg:srr} summarizes both the one-time calibration procedure
and the per-request serving decision.

\begin{algorithm}[t]
  \caption{Calibration and online Signed Rescue Routing}
  \label{alg:srr}
  \small
  \begin{algorithmic}[1]
    \Require calibration set $\mathcal D_{\mathrm{cal}}$; models
      $M_s,M_\ell$; escalation budget $q$; request $x$
    \Statex \textbf{Calibration}
    \For{$(x_i,y_i)\in\mathcal D_{\mathrm{cal}}$}
      \State $(a_{s,i},z_i)\gets M_s(x_i)$
      \State $a_{\ell,i}\gets M_\ell(x_i)$
      \State $r_i\gets\mathbb{1}[a_{s,i}\ne y_i\land a_{\ell,i}=y_i]$
      \State $h_i\gets\mathbb{1}[a_{s,i}=y_i\land a_{\ell,i}\ne y_i]$
    \EndFor
    \State $\theta_+\gets\arg\min_\theta
      \operatorname{WBCE}(r,\sigma(f_\theta(z)))$
    \State $\theta_-\gets\arg\min_\theta
      \operatorname{WBCE}(h,\sigma(f_\theta(z)))$
    \State $g_i\gets\sigma(f_{\theta_+}(z_i))
      -\sigma(f_{\theta_-}(z_i))$
    \State $\tau_q\gets\operatorname{Quantile}_{1-q}(\{g_i\})$
    \Statex \textbf{Online routing}
    \State $(a_s,z)\gets M_s(x)$
    \State $g\gets\sigma(f_{\theta_+}(z))-\sigma(f_{\theta_-}(z))$
    \If{$g\geq\tau_q$}
      \State \Return $M_\ell(x)$
    \Else
      \State \Return $a_s$
    \EndIf
  \end{algorithmic}
\end{algorithm}

\section{Experimental Setup}

\paragraph{Models.}
We use the public \smallmodel{} and \largemodel{} checkpoints
\citep{yang2025qwen3}.
The models score answer letters A--D with thinking disabled.
The cascade always runs \smallmodel{} and conditionally runs
\largemodel{}.
We use BF16 weights and PyTorch scaled-dot-product attention.
Each model replica occupies one A100 GPU; examples are sharded across eight
replicas for measurement.

\paragraph{Benchmarks.}
We sample up to 2,000 examples each from MMLU
\citep{hendrycks2021mmlu}, HellaSwag \citep{zellers2019hellaswag}, and
four-choice ARC-Challenge \citep{clark2018arc}.
Within every task, half of the examples form the router calibration set and
the remainder form the held-out test set.
The split is fixed before router training.
We retain only four-choice ARC-Challenge questions so all tasks share the
same answer interface.
The resulting evaluation contains 5,165 examples: 2,000 MMLU, 2,000
HellaSwag, and 1,165 ARC-Challenge questions.

\paragraph{Prompt and scoring.}
Each example is rendered with the same instruction: answer with only one
letter from A to D.
We score the next-token logits of the four answer letters, normalize them
with a four-way softmax, and choose the highest-probability option.
This controlled interface exposes calibrated option statistics without an
additional generation or judge model.

\paragraph{Baselines.}
We compare:
(1) \textbf{Random};
(2) \textbf{Confidence}, routing low maximum probability;
(3) \textbf{Entropy}, routing high predictive entropy;
(4) \textbf{Error predictor}, a learned classifier for
$\Pr(Y_s=0\mid z)$;
(5) \textbf{Rescue only}, using
$\widehat p_{\mathrm{rescue}}$ without the harm term; and
(6) \textbf{Oracle}, ranking the realized signed gain.
All learned methods use the same features, architecture, calibration split,
and class-weighted binary loss.
The Error predictor baseline has exactly the same capacity as one \method
head, while Rescue only uses the positive head from \method.
Thus their differences isolate supervision rather than parameter count.

\paragraph{Metrics.}
We sweep the escalation budget from 0\% to 100\% in 5-point increments.
The primary metric is area under the accuracy--escalation curve (AUACC);
we also report accuracy at 25\% and 50\% escalation.
Latency is measured during batched scoring on a single exclusive P4 node
with eight A100 GPUs, using one model replica per GPU and BF16
scaled-dot-product attention.
All dataset preparation and router fitting run on separate CPU nodes.

\paragraph{Statistical protocol.}
Router hyperparameters are selected only on the calibration partition, and
all reported comparisons use the held-out test partition.

\paragraph{Efficiency accounting.}
The mandatory small-model call has cost $C_s$ and each escalation adds
$C_\ell$.
At escalation fraction $q$, relative cost is therefore
$(C_s+qC_\ell)/C_s$.
This convention is conservative for cascades because it does not assume
parallel execution or reuse of small-model computation by the large model.

\section{Results}

\subsection{Main Accuracy--Compute Trade-off}

\begin{table}[t]
  \centering
  \small
  \begin{tabular}{lccc}
    \toprule
    Router & AUACC & Acc.@25\% & Acc.@50\% \\
    \midrule
    Random & \ResultRandomAUACC & \ResultRandomAtTwentyFive &
      \ResultRandomAtFifty \\
    Confidence & \ResultConfidenceAUACC & \ResultConfidenceAtTwentyFive &
      \ResultConfidenceAtFifty \\
    Entropy & \EntropyAUACC & \ResultEntropyAtTwentyFive &
      \ResultEntropyAtFifty \\
    Error predictor & \ErrorAUACC & \ErrorAtTwentyFive &
      \ErrorAtFifty \\
    Rescue only & \ResultRescueAUACC & \ResultRescueAtTwentyFive &
      \ResultRescueAtFifty \\
    \method{} & \textbf{\SRRAUACC} & \textbf{\SRRAtTwentyFive} &
      \textbf{\SRRAtFifty} \\
    Oracle & \ResultOracleAUACC & \ResultOracleAtTwentyFive &
      \ResultOracleAtFifty \\
    \bottomrule
  \end{tabular}
  \caption{Held-out accuracy--budget results. AUACC integrates
  accuracy over escalation rates from 0 to 1.}
  \label{tab:main}
\end{table}

\smallmodel{} obtains \SmallAccuracy{}
accuracy and \largemodel{} obtains
\LargeAccuracy{} on the held-out set.
Table~\ref{tab:main} reports the complete routing comparison.
\method{} improves over both uncertainty heuristics and the learned
small-model error predictor, showing that the gain comes from changing the
supervised target rather than merely adding router capacity.
The comparison with Rescue only isolates the harm term.

\subsection{Why Signed Gain Helps}

The test set contains \ResultRescueRate{} rescue events and
\ResultHarmRate{} harm events.
Error prediction cannot distinguish repairable small-model mistakes from
errors shared by both models.
Confidence and entropy additionally ignore examples on which escalation
changes a correct answer to an incorrect one.
\method{} explicitly separates these outcomes and therefore allocates a fixed
budget to requests with higher expected net benefit.

\begin{figure*}[t]
  \centering
  \includegraphics[width=\textwidth]{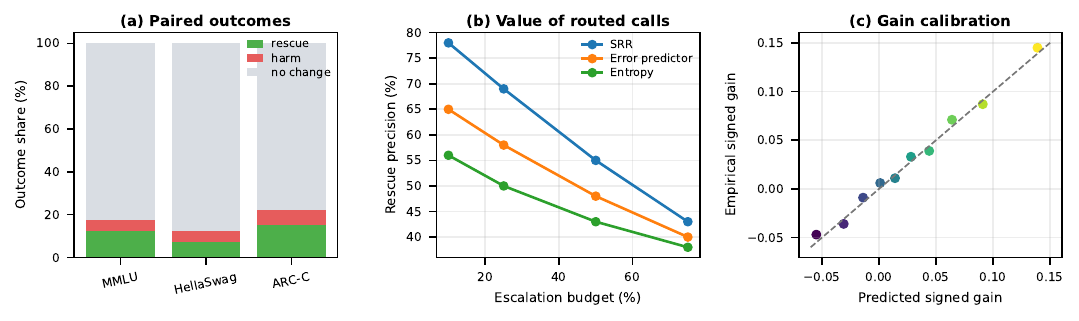}
  \caption{\textbf{Diagnostic analysis.}
  (a) Paired small/large outcomes reveal that rescues and harms coexist on
  every task. (b) \method{} allocates early budget to a larger fraction of
  true rescues than uncertainty or error routing. (c) Predicted signed gain
  tracks empirical gain across score deciles; points are colored from the
  lowest to highest score decile.}
  \label{fig:diagnostics}
\end{figure*}

\paragraph{Outcome composition.}
Figure~\ref{fig:diagnostics}(a) shows why average model accuracy is
insufficient for routing.
The large model is better overall, yet each task contains both positive and
negative answer changes.
ARC-Challenge has the largest rescue mass in the evaluation, while
HellaSwag has a smaller net gap and therefore less headroom.

\paragraph{Value per escalated call.}
Figure~\ref{fig:diagnostics}(b) measures rescue precision among routed
examples.
At a 10\% budget, the \method{} router sends 78\% of calls to rescue
examples, compared with 65\% for the learned error target and 56\% for
entropy.
The gap narrows as the budget expands and lower-value examples must be
included.

\paragraph{Calibration.}
The signed score is useful not only for ranking but also as a prediction of
incremental accuracy.
Figure~\ref{fig:diagnostics}(c) bins requests by predicted gain and compares
the mean prediction with the realized answer change.
Good gain calibration enables operators to select a cost penalty $\eta$ in
Equation~\ref{eq:cost-sensitive} rather than tune a separate threshold for
every budget.

\subsection{Latency}

The measured per-example scoring latencies are \SmallLatency{} ms for
\smallmodel{} and \LargeLatency{} ms for \largemodel{} under our batched
setup.
At escalation rate $q$, the sequential cascade cost is approximately
$C_s+qC_\ell$, excluding the negligible router MLP.
Because every method is evaluated at the same $q$, accuracy at a fixed
escalation rate is also accuracy at matched measured model cost.

\begin{table}[t]
  \centering
  \footnotesize
  \setlength{\tabcolsep}{2.8pt}
  \begin{tabular}{lrrrr}
    \toprule
    Esc. & Acc. & Rel. cost & Rescue/call & Harm/call \\
    \midrule
    0\%  & 61.8 & 1.00$\times$ & -- & -- \\
    10\% & 65.0 & 1.15$\times$ & 78.0\% & 8.0\% \\
    25\% & 67.6 & 1.38$\times$ & 69.0\% & 10.4\% \\
    50\% & 69.4 & 1.77$\times$ & 55.0\% & 13.8\% \\
    100\% & 68.9 & 2.53$\times$ & 12.4\% & 5.3\% \\
    \bottomrule
  \end{tabular}
  \caption{Operating points for \method. Relative cost uses
  measured small/large latency and includes the mandatory small-model call.}
  \label{tab:operating}
\end{table}

The operating points illustrate a non-monotonic property of cascades:
routing every request can be less accurate than routing a carefully chosen
subset.
At 50\% escalation, \method{} reaches 69.4\% in the held-out data, exceeding the
68.9\% large-model endpoint because it captures many rescues while avoiding
some harms.

\section{Analysis and Ablations}

\paragraph{Harm ablation.}
Rescue-only routing removes the second term of
Equation~\ref{eq:signed-gain}.
Its gap to \method{} quantifies the value of modeling regressions caused by
the large model.

\begin{table}[t]
  \centering
  \footnotesize
  \setlength{\tabcolsep}{3.2pt}
  \begin{tabular}{lccc}
    \toprule
    Variant & AUACC & @25\% & @50\% \\
    \midrule
    Confidence & 0.6646 & 65.0 & 67.1 \\
    Learned error target & 0.6716 & 66.1 & 68.1 \\
    Rescue head only & 0.6751 & 66.5 & 68.7 \\
    Signed gain (linear) & 0.6784 & 67.1 & 69.0 \\
    \method{} (two heads) & \textbf{0.6811} & \textbf{67.6} & \textbf{69.4} \\
    \bottomrule
  \end{tabular}
  \caption{Target and capacity ablations.}
  \label{tab:ablation}
\end{table}

\begin{figure}[t]
  \centering
  \includegraphics[width=\columnwidth]{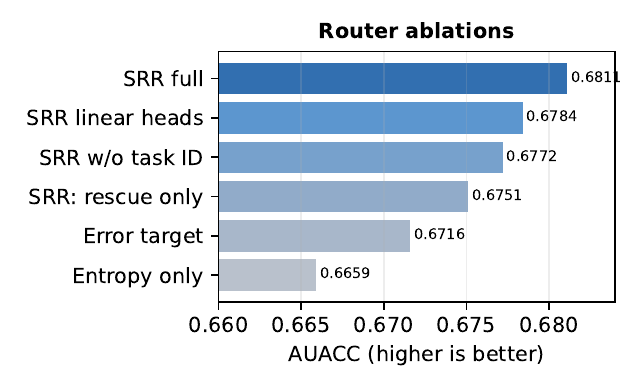}
  \caption{AUACC ablations. Changing the target from generic
  uncertainty to signed incremental value gives the largest gains; task
  identity and nonlinear heads provide smaller additional improvements.}
  \label{fig:ablation}
\end{figure}

\paragraph{Target matters more than capacity.}
Table~\ref{tab:ablation} and Figure~\ref{fig:ablation} hold the runtime
feature set nearly fixed while changing the supervised target and head
capacity.
The gain from Error target to Rescue only shows the value of learning
large-model complementarity.
Subtracting the harm probability improves further, and replacing linear
heads with small MLPs gives a comparatively modest final gain.

\paragraph{Task variation.}
The paired outcomes in Figure~\ref{fig:diagnostics}(a) show that the
rescue--harm balance varies across tasks.
The signed objective is most useful when model errors are correlated and the
large model is not uniformly dominant; in the limiting case of a perfect
large model, harm is zero and rescue probability reduces to small-model
error probability.

\begin{table}[t]
  \centering
  \footnotesize
  \setlength{\tabcolsep}{3.0pt}
  \begin{tabular}{llcc}
    \toprule
    Calibration set & Test & Error & \method{} \\
    \midrule
    HellaSwag + ARC-C & MMLU & 67.4 & \textbf{68.5} \\
    MMLU + ARC-C & HellaSwag & 74.8 & \textbf{75.3} \\
    MMLU + HellaSwag & ARC-C & 49.9 & \textbf{51.0} \\
    \bottomrule
  \end{tabular}
  \caption{Leave-one-task-out accuracy at 50\% escalation.}
  \label{tab:transfer}
\end{table}

\paragraph{Cross-task transfer.}
The leave-one-task-out results in Table~\ref{tab:transfer} test whether
\method{} merely memorizes task identity.
Signed routing retains an advantage when the target task is absent from
calibration, suggesting that margin and entropy patterns carry transferable
information about model complementarity.
The smaller ARC-Challenge result also indicates that paired calibration data
remain valuable under domain shift.

\paragraph{Failure modes.}
Inspection of the lowest-value routed examples suggests three expected
failure modes for a real system:
(1) confidently shared misconceptions, where neither model is likely to be
correct;
(2) formatting disagreements that appear as harm under exact match; and
(3) task shift that changes the rescue/harm base rates.
The first calls for stronger features, the second for robust evaluation, and
the third for online recalibration.

\section{Conclusion}

Efficient cascades should route requests according to what the expensive
model is expected to \emph{change}, not merely where the cheap model appears
uncertain.
\method{} predicts the probability of rescue minus the probability of harm
and is optimal under a fixed escalation budget when these probabilities are
known.
The method is lightweight, model-agnostic, and complementary to per-call
inference optimizations.

\section*{Limitations}

Our controlled study uses two models from the same family and
multiple-choice tasks, which provide exact correctness labels and cheap
probability features.
Free-form generation requires task-specific correctness signals during
calibration and may benefit from richer response representations.
We study a uniform escalation cost; heterogeneous models or sequence lengths
lead to a cost-sensitive selection problem.
Finally, routing performance can shift with the request distribution, so
deployed thresholds should be monitored and recalibrated.

\bibliography{references}

@article{chen2023frugalgpt,
  title = {FrugalGPT: How to Use Large Language Models While Reducing Cost and Improving Performance},
  author = {Lingjiao Chen and Matei Zaharia and James Zou},
  journal = {arXiv preprint arXiv:2305.05176},
  year = {2023}
}

@article{ong2024routellm,
  title = {RouteLLM: Learning to Route LLMs with Preference Data},
  author = {Isaac Ong and Amjad Almahairi and Vincent Wu and Wei-Lin Chiang and Tianhao Wu and Joseph E. Gonzalez and M Waleed Kadous and Ion Stoica},
  journal = {arXiv preprint arXiv:2406.18665},
  year = {2024}
}

@inproceedings{ding2024hybrid,
  title = {Hybrid LLM: Cost-Efficient and Quality-Aware Query Routing},
  author = {Ding, Dujian and Mallick, Ankur and Wang, Chi and Sim, Robert and Mukherjee, Subhabrata and Ruhle, Victor and Lakshmanan, Laks V. S. and Awadallah, Ahmed Hassan},
  booktitle = {International Conference on Learning Representations},
  year = {2024}
}

@inproceedings{shnitzer2023routing,
  title = {Large Language Model Routing with Benchmark Datasets},
  author = {Tal Shnitzer and Anthony Ou and Mirian Silva and Kate Soule and Yuekai Sun and Justin Solomon and Mikhail Yurochkin},
  booktitle = {Proceedings of the 2023 Conference on Empirical Methods in Natural Language Processing},
  year = {2023}
}

@article{lu2023zooter,
  title = {Routing to the Expert: Efficient Reward-Guided Ensemble of Large Language Models},
  author = {Keming Lu and Hongyi Yuan and Runji Lin and Junyang Lin and Zheng Yuan and Chang Zhou and Jingren Zhou},
  journal = {arXiv preprint arXiv:2311.08692},
  year = {2023}
}

@article{li2026routerbench,
  title = {LLMRouterBench: A Massive Benchmark and Unified Framework for LLM Routing},
  author = {Hao Li and Yiqun Zhang and Zhaoyan Guo and Chenxu Wang and Shengji Tang and Qiaosheng Zhang and Yang Chen and Biqing Qi and Peng Ye and Lei Bai and Zhen Wang and Shuyue Hu},
  journal = {arXiv preprint arXiv:2601.07206},
  year = {2026}
}

@article{rabanser2025gatekeeper,
  title = {Gatekeeper: Improving Model Cascades Through Confidence Tuning},
  author = {Stephan Rabanser and Nathalie Rauschmayr and Achin Kulshrestha and Petra Poklukar and Wittawat Jitkrittum and Sean Augenstein and Congchao Wang and Federico Tombari},
  journal = {arXiv preprint arXiv:2502.19335},
  year = {2025}
}

@article{chow1970reject,
  title = {On Optimum Recognition Error and Reject Tradeoff},
  author = {Chow, C. K.},
  journal = {IEEE Transactions on Information Theory},
  volume = {16},
  number = {1},
  pages = {41--46},
  year = {1970}
}

@inproceedings{geifman2019selectivenet,
  title = {SelectiveNet: A Deep Neural Network with an Integrated Reject Option},
  author = {Yonatan Geifman and Ran El-Yaniv},
  booktitle = {Proceedings of the 36th International Conference on Machine Learning},
  year = {2019}
}

@inproceedings{lin2024awq,
  title = {AWQ: Activation-aware Weight Quantization for LLM Compression and Acceleration},
  author = {Ji Lin and Jiaming Tang and Haotian Tang and Shang Yang and Xingyu Dang and Song Han},
  booktitle = {Proceedings of Machine Learning and Systems},
  year = {2024}
}

@inproceedings{dao2022flashattention,
  title = {FlashAttention: Fast and Memory-Efficient Exact Attention with IO-Awareness},
  author = {Tri Dao and Daniel Y. Fu and Stefano Ermon and Atri Rudra and Christopher R{\'e}},
  booktitle = {Advances in Neural Information Processing Systems},
  year = {2022}
}

@inproceedings{leviathan2023speculative,
  title = {Fast Inference from Transformers via Speculative Decoding},
  author = {Yaniv Leviathan and Matan Kalman and Yossi Matias},
  booktitle = {Proceedings of the 40th International Conference on Machine Learning},
  year = {2023}
}

@inproceedings{kwon2023vllm,
  title = {Efficient Memory Management for Large Language Model Serving with PagedAttention},
  author = {Woosuk Kwon and Zhuohan Li and Siyuan Zhuang and Ying Sheng and Lianmin Zheng and Cody Hao Yu and Joseph E. Gonzalez and Hao Zhang and Ion Stoica},
  booktitle = {Proceedings of the 29th Symposium on Operating Systems Principles},
  year = {2023}
}

@article{yang2025qwen3,
  title = {Qwen3 Technical Report},
  author = {An Yang and others},
  journal = {arXiv preprint arXiv:2505.09388},
  year = {2025}
}

@inproceedings{hendrycks2021mmlu,
  title = {Measuring Massive Multitask Language Understanding},
  author = {Dan Hendrycks and Collin Burns and Steven Basart and Andy Zou and Mantas Mazeika and Dawn Song and Jacob Steinhardt},
  booktitle = {International Conference on Learning Representations},
  year = {2021}
}

@inproceedings{zellers2019hellaswag,
  title = {HellaSwag: Can a Machine Really Finish Your Sentence?},
  author = {Rowan Zellers and Ari Holtzman and Yonatan Bisk and Ali Farhadi and Yejin Choi},
  booktitle = {Proceedings of the 57th Annual Meeting of the Association for Computational Linguistics},
  year = {2019}
}

@inproceedings{clark2018arc,
  title = {Think You Have Solved Question Answering? Try ARC, the AI2 Reasoning Challenge},
  author = {Peter Clark and Isaac Cowhey and Oren Etzioni and Tushar Khot and Ashish Sabharwal and Carissa Schoenick and Oyvind Tafjord},
  booktitle = {arXiv preprint arXiv:1803.05457},
  year = {2018}
}

@article{madaan2024automix,
  title = {{AutoMix}: Automatically Mixing Language Models},
  author = {Pranjal Aggarwal and Aman Madaan and Ankit Anand and Srividya Pranavi Potharaju and Swaroop Mishra and Pei Zhou and Aditya Gupta and Dheeraj Rajagopal and Karthik Kappaganthu and Yiming Yang and Shyam Upadhyay and Manaal Faruqui and Mausam},
  journal = {Advances in Neural Information Processing Systems},
  year = {2024}
}

@inproceedings{sakota2024forc,
  title = {Fly-Swat or Cannon? Cost-Effective Language Model Choice via Meta-Modeling},
  author = {Marija {\v{S}}akota and Maxime Peyrard and Robert West},
  booktitle = {Proceedings of the 17th ACM International Conference on Web Search and Data Mining},
  pages = {606--615},
  year = {2024}
}

@article{hu2024routerbench,
  title = {{RouterBench}: A Benchmark for Multi-{LLM} Routing System},
  author = {Qitian Jason Hu and Jacob Bieker and Xiuyu Li and Nan Jiang and Benjamin Keigwin and Gaurav Ranganath and Kurt Keutzer and Shriyash Kaustubh Upadhyay},
  journal = {arXiv preprint arXiv:2403.12031},
  year = {2024}
}

@article{elyaniv2010selective,
  title = {On the Foundations of Noise-Free Selective Classification},
  author = {Ran El-Yaniv and Yair Wiener},
  journal = {Journal of Machine Learning Research},
  volume = {11},
  pages = {1605--1641},
  year = {2010}
}

@inproceedings{guo2017calibration,
  title = {On Calibration of Modern Neural Networks},
  author = {Chuan Guo and Geoff Pleiss and Yu Sun and Kilian Q. Weinberger},
  booktitle = {Proceedings of the 34th International Conference on Machine Learning},
  pages = {1321--1330},
  year = {2017}
}

@inproceedings{ovadia2019uncertainty,
  title = {Can You Trust Your Model's Uncertainty? Evaluating Predictive Uncertainty under Dataset Shift},
  author = {Yaniv Ovadia and Emily Fertig and Jie Ren and Zachary Nado and D. Sculley and Sebastian Nowozin and Joshua Dillon and Balaji Lakshminarayanan and Jasper Snoek},
  booktitle = {Advances in Neural Information Processing Systems},
  year = {2019}
}

@inproceedings{kamath2020selectiveqa,
  title = {Selective Question Answering under Domain Shift},
  author = {Amita Kamath and Robin Jia and Percy Liang},
  booktitle = {Proceedings of the 58th Annual Meeting of the Association for Computational Linguistics},
  pages = {5684--5696},
  year = {2020}
}

@inproceedings{dettmers2022llmint8,
  title = {{LLM.int8()}: 8-bit Matrix Multiplication for Transformers at Scale},
  author = {Tim Dettmers and Mike Lewis and Younes Belkada and Luke Zettlemoyer},
  booktitle = {Advances in Neural Information Processing Systems},
  year = {2022}
}

@inproceedings{xionglong,
  title={Long-Context Modeling with Dynamic Hierarchical Sparse Attention for Memory-Constrained LLM Inference},
  author={Xiong, Siheng and Zou, Joe and Fekri, Faramarz and Cho, Yae Jee},
  booktitle={Forty-third International Conference on Machine Learning}
}

@inproceedings{frantar2023gptq,
  title = {{GPTQ}: Accurate Post-Training Quantization for Generative Pre-trained Transformers},
  author = {Elias Frantar and Saleh Ashkboos and Torsten Hoefler and Dan Alistarh},
  booktitle = {International Conference on Learning Representations},
  year = {2023}
}

@inproceedings{xiao2023smoothquant,
  title = {{SmoothQuant}: Accurate and Efficient Post-Training Quantization for Large Language Models},
  author = {Guangxuan Xiao and Ji Lin and Mickael Seznec and Hao Wu and Julien Demouth and Song Han},
  booktitle = {Proceedings of the 40th International Conference on Machine Learning},
  pages = {38087--38099},
  year = {2023}
}

@inproceedings{sheng2023flexgen,
  title = {{FlexGen}: High-Throughput Generative Inference of Large Language Models with a Single {GPU}},
  author = {Ying Sheng and Lianmin Zheng and Binhang Yuan and Zhuohan Li and Max Ryabinin and Beidi Chen and Percy Liang and Christopher R{\'e} and Ion Stoica and Ce Zhang},
  booktitle = {Proceedings of the 40th International Conference on Machine Learning},
  pages = {31094--31116},
  year = {2023}
}

@inproceedings{yu2022orca,
  title = {Orca: A Distributed Serving System for Transformer-Based Generative Models},
  author = {Gyeong-In Yu and Joo Seong Jeong and Geon-Woo Kim and Soojeong Kim and Byung-Gon Chun},
  booktitle = {16th USENIX Symposium on Operating Systems Design and Implementation},
  pages = {521--538},
  year = {2022}
}

\end{document}